\documentclass[letterpaper, 10 pt, conference]{ieeeconf}  

\IEEEoverridecommandlockouts                              

\usepackage{amsmath,epsfig}
\usepackage{multirow}

\usepackage[natbib, bibencoding=utf8, citestyle=numeric, bibstyle=ieee, maxbibnames=999, maxcitenames=2, mincitenames=1, sortcites]{biblatex}

\bibliography{refs.bib}

\usepackage[capitalise, nameinlink, noabbrev]{cleveref}

\title{\LARGE \bf
Journaling Data for Daily PHQ-2 Depression Prediction and Forecasting
}

\author{Alexander Kathan$^{1*}$, Andreas Triantafyllopoulos$^{1}$, Xiangheng He$^{1,2}$, Manuel Milling$^{1}$, Tianhao Yan$^{1}$,\\Srividya Tirunellai Rajamani$^{1}$, Ludwig Küster$^{3}$, Mathias Harrer$^{3,4}$, Elena Heber$^{3}$, Inga Grossmann$^{3}$,\\ David D. Ebert$^{3,4}$, Björn W.\ Schuller$^{1,2}$
\thanks{$^{1}$A.\,K., A.\,T., X.\,H., M.\,M., T.\,Y., S.\,T.\,R., and B.\,S. are with the Chair of Embedded Intelligence for Health Care \& Wellbeing, University of Augsburg, Germany}%
\thanks{$^{2}$B.\,S. and X.\,H. are also with GLAM -- the Group on Language, Audio,
\& Music, Imperial College London, London, UK}
\thanks{$^{3}$L.\,K., M.\,H., E.\,H., I.\,G. and D.\,E. are with GET.ON Institut für Online Gesundheitstrainings GmbH/HelloBetter, Hamburg, Germany}
\thanks{$^{4}$M.\,H., and D.\,E. are also with the Chair of Psychology \& Digital Mental Health Care, Technical University Munich, Germany}
\thanks{$^{*}$ Corresponding author:
        {\tt\small alexander.kathan@uni-a.de}}%
\thanks{© 2022 IEEE.  Personal use of this material is permitted.  Permission from IEEE must be obtained for all other uses, in any current or future media, including reprinting/republishing this material for advertising or promotional purposes, creating new collective works, for resale or redistribution to servers or lists, or reuse of any copyrighted component of this work in other works.}
}

\begin{document}\sloppy

\maketitle
\thispagestyle{empty}
\pagestyle{empty}

\begin{abstract}
Digital health applications are becoming increasingly important for assessing and monitoring the wellbeing of people suffering from mental health conditions like depression. 
A common target of said applications is to predict the results of self-assessed Patient-Health-Questionnaires (PHQ), indicating current symptom severity of depressive individuals.
Many of the currently available approaches to predict PHQ scores use passive data, e.\,g., from smartphones. 
However, there are several other scores and data besides PHQ, e.\,g., the Behavioral Activation for Depression Scale–Short Form (BADSSF), the Center for Epidemiologic Studies Depression Scale (CESD), or the Personality Dynamics Diary (PDD), all of which can be effortlessly collected on a daily basis.
In this work, we explore the potential of using actively-collected data to predict and forecast daily PHQ-2 scores on a newly-collected longitudinal dataset.
We obtain a best MAE of $1.417$ for daily prediction of PHQ-2 scores, which specifically in the used dataset have a range of $0$ to $12$,
using leave-one-subject-out cross-validation, as well as a best MAE of $1.914$ for forecasting PHQ-2 scores using data from up to the last $7$ days.
This illustrates the additive value that can be obtained by incorporating actively-collected data in a depression monitoring application.

\end{abstract}

\section{Introduction}

Depression is a common mental illness that affects millions of individuals around the world, leading to a variety of emotional and physical problems that pose a threat to a patient's overall wellbeing, and in many cases the demand for help exceeds available resources, e.\,g., among college students~\citep{beck2014depression,auerbach2016mental}. 
As its timely diagnosis and longitudinal monitoring can inform treatment decisions, recent years have seen a rise in (digital) monitoring applications~\citep{sequeira2020mobile}.
These applications may collect various data streams such as movement~\citep{sequeira2020mobile}, heart rate signals~\citep{wang2018tracking}, or speech~\citep{cummins2018speech}.

Furthermore, it is common to employ self-administered questionnaires for assessing the severity of depression symptoms, such as the 2- and 9-item Patient-Health-Questionnaires (PHQ-2/9)~\citep{kroenke2003patient, lowe2004measuring}.
These questionnaires are well-correlated with symptom severity, amount of sick days, and higher healthcare utilisation, and thus form well-validated prediction and forecasting targets for digital health applications.

Predicting those depression severity scales is usually done with two kinds of data: passively- and actively-collected. 
In recent years, there has been a strong focus on passive data by using a broad range of wearable devices~\cite{han2021deep, qian2021artificial}, which have been used to detect not only depression, but also a number of other diseases such as COVID-19~\cite{liu2022fitbeat}. 

The recent review of \citet{sequeira2020mobile} analysed a range of mobile and wearable technology for monitoring of depressive symptoms and came to the conclusion that certain mobile technologies are able to track depression.
In line with this conclusion, \citet{lu2018joint} use passive phone data and combine them in a multi-task learning model for depression assessment. 
\citet{pratap2019accuracy} also use passively-collected phone data to predict daily mood. 
Within their study, they found out that for predicting daily PHQ-2, passive phone data and their derived features may not be suited for a prediction at a population level. 
This is caused by the high variation in phone usage patterns and daily mood ratings. 
\citet{ringeval2019avec} chose a different approach to depression prediction by taking audio and video as available data streams into account~\cite{ringeval2019avec}. 

Different from previous work, our study focuses on exclusively exploring active data in the form of diary questionnaires that can be effortlessly collected on a daily basis for PHQ-2 prediction and forecasting at a population level. These kinds of active assessments are known as Experience Sampling Methods (ESM)~\cite{rhee2020experience} and are often used for journaling to track symptoms. Symptom monitoring is a typical part of cognitive-behavioural treatments for depression~\cite{telford2012experience}, as well as for other mental disorders~\cite{vachon2019compliance}. Therefore, this type of assessment is common and often used in practice. However, especially under real-world conditions, these data are not typically used for modelling, thus their potential is not yet fully exploited, so we intend to explore these possibilities within this work by predicting and forecasting PHQ-2 
ratings 
with it.
In doing so, we consider not only individual questions in isolation, but a number of different scores that participants record daily in a smartphone/app diary. 

The remainder of this paper is organised as follows.
\cref{sec:maikidataset} describes the MAIKI study and the resulting dataset.
\cref{sec:experimentalsetup} and \cref{sec:experiments} outline our experiments and results.
\cref{sec:conclusion} concludes the paper with a brief discussion.

\section{MAIKI Dataset}
\label{sec:maikidataset}
In this section, we provide information about the longitudinal MAIKI dataset, a subset of which we use in this work. It is collected within the ``Mobile daily living therapy assistant with interaction-focused artificial intelligence for depression'' (MAIKI) project, in which various active and passive data modalities, e.\,g., phone data, GPS data, or several questionnaires were gathered from 48 patients over a period of three months. The study procedures were approved by the ethics committee of the Friedrich-Alexander-University Erlangen-Nuremberg (385\_20B).

MAIKI is a real-world dataset. During the period of the study, there have been days when no data was collected from some participants. The reason for this is that some individuals did not complete the diaries for everyday. In \cref{sec:experimentalsetup}, we explain in detail how we handle missing values to ensure data quality.
The various data streams available from all participants of the study are outlined below. 

\subsection{Passively-collected data}
First, passive phone data were collected. These can be roughly divided into the four areas of (1) app sessions \& app usage information, (2) metadata on general cell phone settings \& phone actions, (3) GPS data, as well as (4) communication information in the form of contacts, call \& SMS data. 

Since this work focuses on the actively-collected daily data, we will not go into further in-depth for passive data, but instead describe the active data in more detail.


\subsection{Actively-collected data}
In parallel to the passively-collected data, the project also collected active data. Active data collection took place in four different ways: (1) First, audio data were recorded at the beginning of the study. This was done during an interview with a trained psychologist. (2) Second, a self-assessment survey was collected at least once from each participant, providing different socio-demographic information. In addition, these surveys determined various clinical scores such as the Assessment of Quality of Life (AQoL-8D)~\cite{richardson2014validity}, Behavioral Activation for Depression Scale-Short Form (BADSSF)~\cite{manos2011behavioral}, or the Big Five Inventory (BFI)~\cite{john1991big}. (3) Third, each patient underwent a weekly screening in which the three scores Generalized Anxiety Disorder (GAD-7)~\cite{spitzer2006brief}, Perceived Stress Scale (PSS-4)~\cite{lee2012review}, and Patient Health Questionnaire (PHQ-9)~\cite{lowe2004measuring} were collected. (4) Fourth, study participants kept a diary in which they answered questions daily. These questions were based on items extracted from psychometrically validated questionnaires: the Center for Epidemiologic Studies Depression (CESD; item 5, 7, 20)~\cite{lewinsohn1997center}, Personality Dynamics Diary (PDD; agentic/communal reward, workload)~\cite{zimmermann2019integrating}, Pittsburgh Sleep Quality Index (PSQI; item 6)~\cite{buysse1989pittsburgh}, BADS-SF (item 1, 5, 7), as well as PHQ-2~\cite{kroenke2003patient}. 
Different to previous studies, a scale of 0-12 is used for PHQ-2 in the MAIKI dataset, aiming at an even finer patient state estimation. A division by 2 allows a conversion to the commonly used PHQ-2 scale, ranging from 0-6.
In addition to the diary, data was also collected during the day using the ESM, providing information about the current mood of the individuals.

\subsection{Feature sets}
From all the assembled data in the MAIKI dataset, we use the actively-collected daily data which was gathered by a diary entry of each participant for every day in the study. Based on the diaries data, we compiled three feature sets as follows:
\begin{itemize}
    \item \textbf{Intraday-ESM} data contains only four features that reflect a person's daily mood, including scores for activity, happiness, sadness, and tension during the day. In addition, it contains a PSQI score, indicating the quality of sleep of the previous night.
    \item \textbf{Daily-Diary} consists of questions of the depression scores BADSSF and CESD, as well as PDD. Further, it includes a score for perceived anxiety and experiential avoidance.
    \item \textbf{Intraday-ESM \& Daily-Diary} combines both feature sets and therefore contains a broad spectrum of information about the daily mood of a patient.
\end{itemize}
\section{Experimental setup}
\label{sec:experimentalsetup}
In this section, we provide details on the experimental setup.
All models were evaluated in a Leave-One-Subject-Out (LOSO) cross-validation setup, where for each fold, we test the model on one participant and train on all the others. Optimisation was performed by nested-cross-validation on each train data set. 
Features are min-max normalised to a $[0, 1]$ range for each fold.
We also experimented with mean-std normalisation, but do not report results on it, as it showed inferior performance on our dataset.
As an initial baseline for the MAIKI dataset, we calculated the mean absolute error (MAE) based on the rolling-mean value of all labels for each subject, which corresponds to chance-level. Due to the fact that features from Intraday-ESM and Daily-Diary are not always both available on all days, there are separate baselines for the individual two feature sets, since a different number of labels corresponding to the features leads to another mean value.

In the following, we explain which method and strategy we use to deal with missing data in all feature sets in order to ensure data quality. 
Additionally, we describe the different models we use to perform PHQ-2 prediction and forecasting based on the actively-collected ESM and diaries. 

\subsection{Missing data}
As described in section \ref{sec:maikidataset}, MAIKI is a real-world dataset and therefore also has missing data, for which we apply the following strategy. 

At first, we filter the data by considering a participant's entry for each associated label as available only if they have provided at least five days of data in the week prior to the corresponding label date. The data missing rate of the subsequent filtered dataset comprises 16\,\% on average for the three feature sets (17\,\% in Intraday-ESM and the combined feature set and 14\,\% in Daily-Diary).

For dealing with this missing data, we perform standard statistical imputation. At every missing data point, we perform linear interpolation by calculating the mean value between the last and the next available data point. Therefore, for each missing feature of one day, the mean feature value of the previous day and the following day is calculated.

\subsection{Models}
We experiment with four different regression models: a) XGBoost~\cite{chen2016xgboost}, b) Support Vector Machines (SVMs), c) Random Forests, and d) Multilayer Perceptrons (MLPs).
Each model comes with distinct associated hyperparameters which we tune using 3-fold nested-cross-validation on each train data set with non-disjoint subjects. 
For XGBoost, we optimise the column subsample ratio (\{$0.2, 0.4, 0.6, 0.8\}$), the maximum tree depth ($\{3, 4, 5\}$), and the number of trees trained ($\{10, 100\}$).
For SVM, we optimise the kernel function ({\textit{RBF}, \textit{linear}}) and the $ cost (C)$ value ($\{0.0001, 0.001, 0.1, 1, 3, 5, 10\}$). 
The hyperparameters for Random Forests are the number of trees ($\{10, 100, 500, 1000\}$) and the number of features to consider when looking for the best split ($\{2, 3, 4, 5, 6\}$).

Finally, for MLPs, we use a fixed architecture of two linear layers with dropout of $0.2$ and ReLU activation function, as well as Adam as an optimiser. The number of neurons in each case corresponds to the number of available features ($\{11, 13, 24\}$) and one neuron for the output layer.  
The models were trained for $50$ epochs with a batch size of $16$, a learning rate of $0.0005$, and MAE Loss.

\section{Results}
\label{sec:experiments}
In this section, we describe the experiments performed, as well as the results achieved. First, we explore the performance of PHQ-2 prediction with data from the same day. Second, we present results for PHQ-2 forecasting for a given day based on the data of up to seven previous days.

\subsection{PHQ-2 prediction with data from the same day}

\begin{table}[t!]
  \begin{center}
  \caption{PHQ-2 (scale: $[0-12]$) prediction\\based on the (active) data from the same day.}
  \vspace{.5em}
  \begin{tabular}{|l|l|l|l|}
    \hline
    \textbf{Data} &  \textbf{Model} &  \textbf{MAE} \\
    \hline
    \hline
    \multirow{5}*{Intraday-ESM} & Baseline & 2.407 \\
    \cline{2-3}
      & XGBoost & 1.972 \\
    \cline{2-3}
      & SVM & 1.892 \\
    \cline{2-3}
      & Random Forest & 1.928 \\
    \cline{2-3}
      & MLP & \textbf{1.855} \\
    \hline
    \hline
    \multirow{5}*{Daily-Diary} & Baseline & 2.499 \\
    \cline{2-3}
      & XGBoost & 1.517 \\
    \cline{2-3}
      & SVM & \textbf{1.455} \\
    \cline{2-3}
      & Random Forest & 1.536 \\
    \cline{2-3}
      & MLP & 1.468 \\
    \hline
    \hline
    \multirow{5}*{Intraday-ESM \& Daily-Diary}  & Baseline & 2.407 \\
    \cline{2-3}
      & XGBoost & 1.617 \\
    \cline{2-3}
      & SVM & 1.496 \\
    \cline{2-3}
      & Random Forest & 1.577 \\
    \cline{2-3}
      & MLP & \textbf{1.417} \\
    \hline
   \end{tabular}
  \label{Results_Daily_PHQ2-Precition}
  \end{center}
\end{table}

Table \ref{Results_Daily_PHQ2-Precition} summarises the regression performance for daily PHQ-2 prediction with active diaries data from the same day which we report with the MAE for each feature set. Furthermore, the table shows an initial baseline MAE that was calculated based on the PHQ-2 rolling-mean value of all participants within the feature set. PHQ-2 ranges from values between 0-12. This range was defined specifically for MAIKI data collection to provide an even finer scale than the regular PHQ-2 range.
In Intraday-ESM and the combined feature set, the best regression performance can be observed with the MLP model, which yields an MAE of 1.855 and 1.417 with a baseline of 2.407. In Daily-Diary, the best result can be observed with the SVM model, which has an MAE of 1.455 with a baseline of 2.499.

\subsection{PHQ-2 forecasting with data from the last 7 days}

\begin{table}[t!]
  \begin{center}
  \caption{PHQ-2 (scale: $[0-12]$) forecasting\\based on the last 1-7 days (active) data.}
  \vspace{.5em}
  \begin{tabular}{|l|l|l|l|}
    \hline
    \textbf{Data} &  \textbf{Model} &  \textbf{MAE} \\
    \hline
    \hline
    \multirow{5}*{Intraday-ESM (5 days)} & Baseline & 2.176 \\
    \cline{2-3}
      & XGBoost & 2.161 \\
    \cline{2-3}
      & SVM & \textbf{2.018} \\
    \cline{2-3}
      & Random Forest & 2.034 \\
    \cline{2-3}
      & MLP & 2.133 \\
    \hline
    \hline
    \multirow{5}*{Daily-Diary (2 days)}  & Baseline & 2.438 \\
    \cline{2-3}
      & XGBoost & 2.185 \\
    \cline{2-3}
      & SVM & 2.110 \\
    \cline{2-3}
      & Random Forest & 2.174 \\
    \cline{2-3}
      & MLP & \textbf{2.048} \\
    \hline
    \hline
    \multirow{5}*{Intraday-ESM \& Daily-Diary (4 days)}  & Baseline & 2.176 \\
    \cline{2-3}
      & XGBoost & 1.967 \\
    \cline{2-3}
      & SVM & 1.964 \\
    \cline{2-3}
      & Random Forest & 2.022 \\
    \cline{2-3}
      & MLP & \textbf{1.914} \\
    \hline
   \end{tabular}
  \label{Results_Tomorrows_PHQ2_Prediction}
  \end{center}
\end{table}

\cref{Results_Tomorrows_PHQ2_Prediction} summarises for each feature set the best regression performance of PHQ-2 forecasting for the next day with data from up to the last 1-7 days. For Intraday-ESM, the best result is achieved with the SVM model with data from the last 5 days, which leads to an MAE of 2.018 with a baseline of 2.176. For Daily-Diary, the best result can be observed with the MLP model, which yields an MAE of 2.048 with a baseline of 2.438. For Intraday-ESM \& Daily-Diary, the MLP model leads to the best result with an MAE of 1.914 and a baseline of 2.176. 

\begin{figure}
  \includegraphics[width=.485\textwidth]{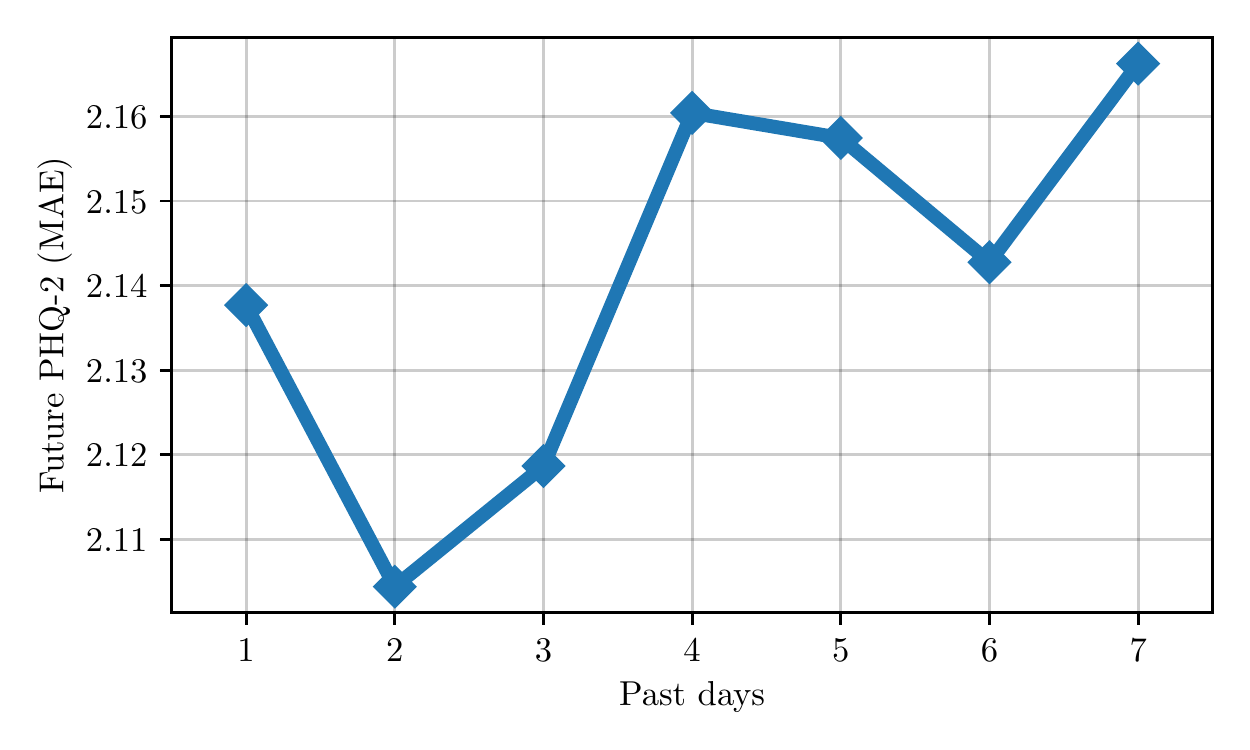}
  \caption{
  PHQ-2 (scale: $[0-12]$ future score prediction using data from the last $[1-7]$ days for the MLP classifier with Daily-Diary.
  }
  \label{fig:mlp_days}
\end{figure}

\cref{fig:mlp_days} shows the variation of performance when adding further past days ($[1-7]$). 
This demonstrates that more days do not lead to better results in the case of PHQ-2 forecasting using active diaries and ESM data. 
In fact, minor fluctuations can be observed, which are, however, not statistically significant (t-tests for MAE values when using only previous day vs all other days, $\alpha=0.05$). 
A similar behaviour can be observed for all three feature sets. 

\section{Conclusion and future work}
\label{sec:conclusion}


For early detection, predicting scores for depression detection is becoming increasingly important in order to be able to intervene in time. With the normal PHQ-2 scale ($[0-6]$), intervention should take place from a score of 3, which corresponds to a score of 6 on our chosen scale ($[0-12]$). We showed that PHQ-2 can be predicted with a small amount of daily active data. We also observed that the previous day is most important for PHQ-2 forecasting, and that days previous to the last day have only a small effect on performance.

The results are therefore relevant for future depression detection and lead to the conclusion that actively-collected data can be promising to improve prediction models for PHQ scores in the future.

Overall, MLP models had the best performance. Our results demonstrated that active data should be given more consideration in prediction models in the future, e.\,g., by combining more active data with commonly used passive prediction models. 

Future work could be targeted at exploring these different possible multimodal combinations to further improve performance, as well as investigating more complex models.

\section*{Acknowledgments}
\noindent
Data analysed in this publication were collected as part of the MAIKI project, which was funded by the German Federal Ministry of Education and Research (grant No.\ 13GW0254). The responsibility for the content of this publication lies with the authors.


\section*{\refname}
\printbibliography[heading=none]


\end{document}